%% file: main.tex
\documentclass[10pt,twocolumn,letterpaper]{article}

\usepackage{iccv}
\usepackage{times}
\usepackage{epsfig}
\usepackage{graphicx}
\usepackage{amsmath}
\usepackage{amssymb}
\usepackage{overpic}
\usepackage{float}
\usepackage{subfig}
\usepackage[ruled]{algorithm2e}
\usepackage[accsupp]{axessibility}

\usepackage[pagebackref=true,breaklinks=true,letterpaper=true,colorlinks,bookmarks=false]{hyperref}

\iccvfinalcopy 

\def\iccvPaperID{4628} 

\ificcvfinal\fi

\begin{document}

 \title{NeTO:{Ne}ural Reconstruction of {T}ransparent {O}bjects with  \\ Self-Occlusion Aware Refraction-Tracing}


\author{Zongcheng Li$^{1}$$^{\ast}$  \quad Xiaoxiao Long$^{2}$$^{\ast}$ \quad Yusen Wang$^{1}$ \quad Tuo Cao$^{1}$ \\ \quad Wenping Wang$^{3}$ \quad Fei Luo$^{1}$ \quad Chunxia Xiao$^{1}$$^{\dagger}$ \\[0.3em]\\
 \quad $^{1}$School of Computer Science, Wuhan University  \\
 \quad $^{2}$The University of Hong Kong   
\quad $^{3}$Texas A\&M University }

\maketitle

\footnotetext[1]{Equal contributions.}
\footnotetext[2]{Corresponding author.}

\ificcvfinal\thispagestyle{empty}\fi

\begin{abstract}

    We present a novel method called NeTO, for capturing the 3D geometry of solid transparent objects from 2D images via volume rendering. 
    Reconstructing transparent objects is a very challenging task, which is ill-suited for general-purpose reconstruction techniques due to the specular light transport phenomena.
    Although existing refraction-tracing-based methods, designed especially for this task, achieve impressive results, they still suffer from unstable optimization and loss of fine details since the explicit surface representation they adopted is difficult to be optimized, and the self-occlusion problem is ignored for refraction-tracing.
    In this paper, we propose to leverage implicit Signed Distance Function (SDF) as surface representation and optimize the SDF field via volume rendering with a self-occlusion aware refractive ray tracing. 
    The implicit representation enables our method to be capable of reconstructing high-quality reconstruction even with a limited set of views, and the self-occlusion aware strategy makes it possible for our method to accurately reconstruct the self-occluded regions. 
    Experiments show that our method achieves faithful reconstruction results and outperforms prior works by a large margin.
    Visit our project page at \url{https://www.xxlong.site/NeTO/}.
\end{abstract}


	
\input{NeTO/sections/1_intro}

\input{NeTO/sections/2_related_works}

\input{NeTO/sections/3_method}
\input{NeTO/sections/4_implementation}

\input{NeTO/sections/5_experiments}

\input{NeTO/sections/6_conclusion}

{\small
    \bibliographystyle{ieee_fullname}
    \bibliography{egbib}
}

\end{document}

%% file: NeTO/sections/1_intro.tex
\section{Introduction}
\label{sec:intro}

\begin{figure}[tp!]
\centering
\begin{overpic}
[width=1.0\linewidth]{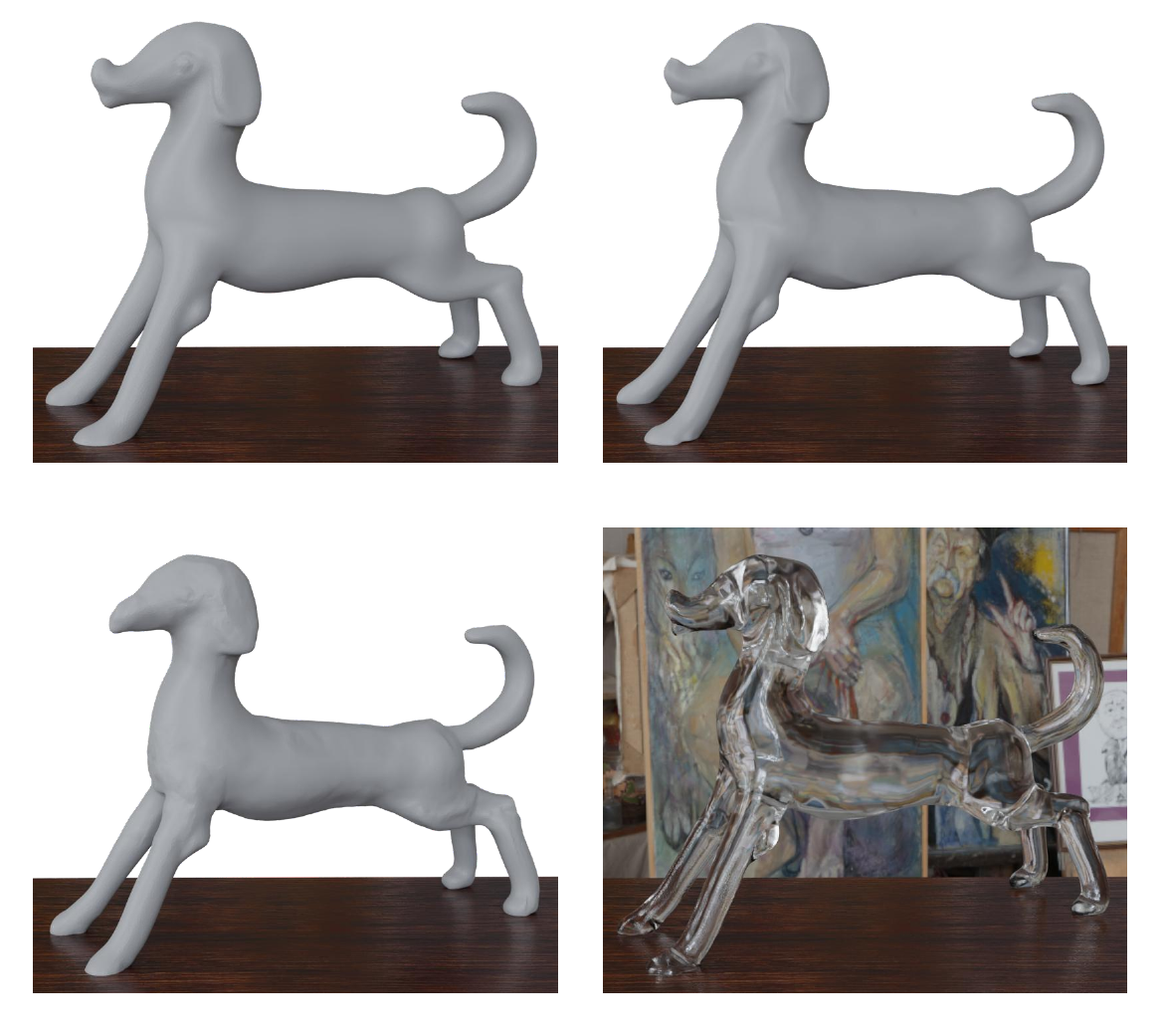}
\put(15, 45.3){\small  Ground Truth}
\put(19, -1.7){\small  DRT~\cite{lyu2020differentiable}}
\put(72,45.3){\small  Ours}
\put(59,-1.7){\small  Our synthesized view }
\end{overpic}
\caption{Illustration of a sparse setting using only one
fourth of the camera images, i.e., $\{I_i \}_{i=1,5,9...}$, to recover the
model Dog in the DRT dataset.
Compared with DRT, our method produces more accurate renderings, which indicates the high quality of our reconstruction. The synthesized view is obtained via Blender.}

\label{figure1}
\end{figure}

\begin{figure*}[t]
\centering
\begin{overpic}
[width=1.0\linewidth]{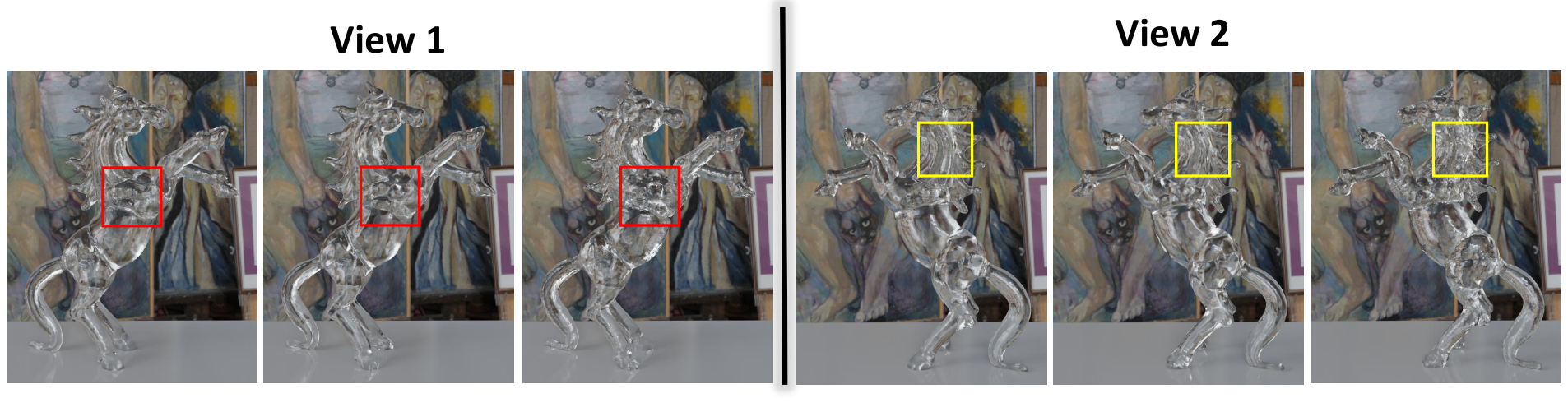}
\put(2.5,-0.7){ Ground Truth}
\put(22,-0.7){ Ours}
\put(37,-0.7){ DRT~\cite{lyu2020differentiable}}
\put(53,-0.7){ Ground Truth}
\put(73,-0.7){ Ours}
\put(87.7,-0.7){ DRT~\cite{lyu2020differentiable}}
\end{overpic}
\caption{The comparisons of novel view synthesis with $sparsity=8$ (9 views).
After obtaining the reconstruction models via our method and DRT, we render two views of the models via Blender.
Compared with DRT, our method produces more accurate renderings (see the red and yellow boxes), which indicates the high quality of our reconstruction.
}

\label{transparent}
\end{figure*}
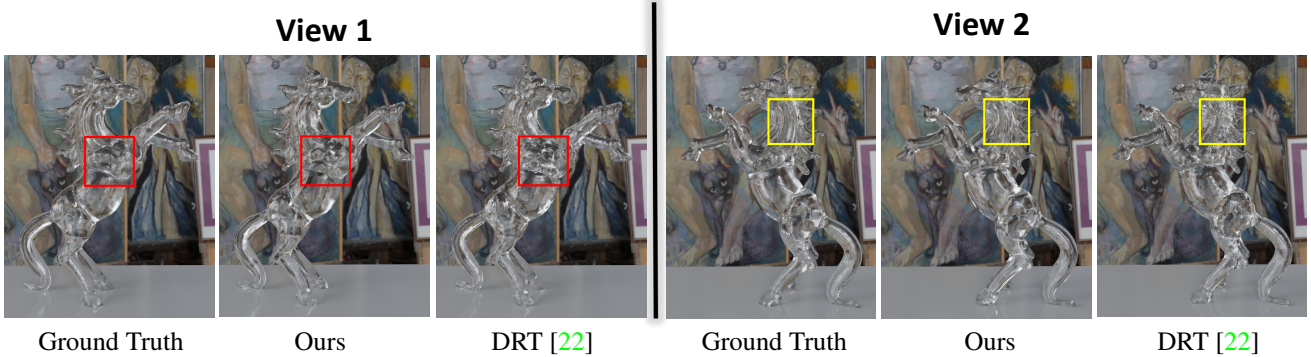

Reconstructing 3D models of real-world objects has been one of the longstanding problems. It has been researched for decades in computer vision and graphics, which boosts the development of many applications, such as augmented reality, automatic driving, and robots. However, existing general-purpose multi-view reconstruction methods~\cite{liao2021adaptive,liao2021dense, schoenberger2016mvs, wang2021neus} are only suitable for opaque objects whose surfaces are approximately Lambertian, and few of them can tackle transparent objects.
The light paths passing through transparent objects are extremely complex and involve refractions and reflections.

Recently, some state-of-the-art methods have been proposed to reconstruct solid transparent objects in a non-intrusive manner, capturing refraction-tracing consistency with specially designed hardware systems, and have produced impressive results. This is achieved by optimizing correspondences between camera rays and locations on the background monitor~\cite{lyu2020differentiable} or enforcing consistency between camera rays and refracted rays with a rotating background monitor~\cite{wu2018full}.
However, those methods either adopt point cloud~\cite{wu2018full} or mesh~\cite{lyu2020differentiable} as surface representation, and the explicit representations are difficult to be optimized. As a result, the methods usually require a large number of views as input for optimization. Without enough images as input, the methods easily fail to reconstruct faithful geometry due to unstable optimization (see Figure~\ref{figure1}).
		
More importantly, a critical issue still remains ignored, i.e., how to tackle the self-occluded parts of the objects.
The widely-used refraction-tracing consistency assumes that a camera ray is only refracted twice (upon entering and upon exiting) on the object surfaces when the ray passes through a transparent object.
However, the assumption is not always true when a camera ray passes through the self-occluded parts where the ray will be refracted by surfaces more than twice.
As a result, mistakenly enforcing the refraction-tracing consistency on the self-occluded parts will unavoidably introduce errors in the optimization of reconstruction, which is a bottleneck to further enhance the reconstructed geometries.

In this work, we propose a novel method, called NeTO, for reconstructing high-quality 3D geometry of solid transparent objects. In contrast to prior works~\cite{lyu2020differentiable}, we adopt implicit Signed Distance Function (SDF) as surface representation and leverage volume rendering~\cite{wang2021neus} to enforce the refraction-tracing consistency. 
Moreover, we propose a simple but effective strategy to detect the self-occluded parts and avoid mistakenly enforcing constraints on these regions.
The key idea is that we leverage the \textbf{law of reversibility}, that is, \textit{If the direction of a light beam is reversed, despite the number of times the beam is reflected or refracted, it will follow the same path}, to identify whether a camera ray is reversible or not upon the assumption that the ray is refracted exactly twice.

To validate our method, we conduct experiments on DRT~\cite{lyu2020differentiable} dataset and our collected data with full views setting and various sparse views settings.
The sparse setting selects one view from every n consecutive camera index, i.e., $\{1, n+1, 2n+1, ...  \}$, where $n$ is termed as \textit{Sparsity}.
The extensive experiments show that our method enables the high-quality reconstruction of transparent objects and outperforms the previous methods.
Our contributions can be summarized as follows:

\begin{itemize}
\item A novel neural surface reconstruction system is adopting implicit SDF as a representation for reconstructing transparent objects, thus enabling robust reconstruction optimization.
\item A self-occlusion aware refraction-tracing strategy is introduced to accurately enforce the constraint, making it possible to recover geometries with fine details.
\item Experimental results show that our method achieves SOTA results compared to prior works.
\end{itemize}

%% file: NeTO/sections/2_related_works.tex
\section{Related Work}
\label{sec:RELATED WORKS}
\subsection{Environment matting}
\label{Environment_matting}
Environment matting is introduced by~\cite{Zongker:1999:EM}, which extracts the environment matte and silhouettes from a series of projected horizontal and vertical stripe patterns. Subsequent works have been extended to multiple cameras~\cite{EME:Extensions}, natural images~\cite{chen2018tom,imageBasedEM}, wavelet domains~\cite{WaveletEM}, and frequency domain~\cite{FrequencyEM}. Meanwhile, it can be combined with compressive sensing theory~\cite{csEM} to reduce the number of used images. 
Unlike the above method, we adopt environment matting to capture environment matte and object masks and optimize the object geometries to fit them.
 
\subsection{Transparent Object Reconstruction}

Recovering the 3D geometry of transparent objects is a longstanding challenging problem~\cite{ihrke2010transparent}. 
To solve this difficult task, many works leverage specially designed hardware setups to provide more information encoding object geometries, including polarization~\cite{cui2017polarimetric, huynh2010shape,miyazaki2005inverse,shao2022polarimetric}, tomography~\cite{trifonov2006tomographic}, a moving point light source~\cite{chen2006mesostructure,morris2007reconstructing}, light field probes~\cite{wetzstein2011refractive} and gray-coded patterns~\cite{lyu2020differentiable,qian20163d,wu2018full}. 
Some methods~\cite{shan2012refractive,yue2014poisson} target the reconstruction of transparent objects with refractive or mirror-like surfaces.
Other methods, including ours, focus on solid transparent objects where most camera rays will refract on the surfaces twice. 
To reconstruct the geometry of transparent objects, there are many types of correspondences proposed, like multi-view ray-ray correspondences~\cite{qian20163d,rayRevealShape,wu2018full}, and ray-location correspondences~\cite{  lyu2020differentiable}. 
DRT~\cite{lyu2020differentiable} proposed to extract per-view ray-location correspondences by using the EnvMatt algorithm~\cite{Zongker:1999:EM}, and utilize the differentiable rendering
for progressively optimizing explicit meshes. 
Xu~\etal~\cite{xu2022hybrid} introduced ray-cell correspondence for reconstructing the full mode under natural light. Besides, Shao~\etal~{~\cite{shao2022polarimetric}} adopted polarimetric cues to reconstruct the full model of transparent objects~\cite{shao2022polarimetric}. 
    
Recently, data-driven-based methods have shown remarkable achievements in estimating the depth and normal maps of transparent objects~\cite{li2020through,sajjan2020clear,stets2019single}.
Li~\etal~\cite{li2020through} first predicted the rough geometry of the transparent objects and then leveraged Pointnet++~\cite{qi2017pointnet++} to further refine the rough geometry.
However, due to the domain gap between the synthetic and real data, Li~\etal~\cite{li2020through} failed to reconstruct real objects unseen in its training dataset.
More recently,  Bemana~\etal~\cite{bemana2022eikonal} leveraged NeRF for novel view synthesis of transparent objects and showed good performance to render novel views. However, since it targets novel view synthesis rather than reconstruction, it's difficult to extract reliable geometry from the method.
Unlike the above methods, We leverage volume rendering to simulate the refraction-tracing path for geometry optimization.
	
\subsection{Neural Implicit Representation}
\label{NIR}
Existing 3D representations can be roughly divided into four categories$:$ voxel-based representations~\cite{choy20163d}, point-based representations~\cite{achlioptas2018learning,fan2017point}, mesh-based representations~\cite{lyu2020differentiable,wang2018pixel2mesh}, and neural implicit representations~\cite{chen2019learning,mescheder2019occupancy,park2019deepsdf,saito2019pifu}. 
Recently, implicit neural representations have been applied to a variety of applications, including novel view synthesis~\cite{mildenhall2020nerf,zhang2020nerf++}, camera pose estimation~\cite{lin2021barf,wang2021nerf}, human~\cite{liu2021neural,peng2021neural} and multi-view 3D reconstruction~\cite{guo2022neural,long2022neuraludf,long2022sparseneus,niemeyer2020differentiable,oechsle2021unisurf,sun2022neural,wang2021neus,wang2022neuralroom,yariv2021volume,yariv2020multiview}, and achieved impressive successes. 
Recent research has shown that reconstructed results using implicit neural representation often produce higher quality than other 3D representations. 

For the task of 3D reconstruction from 2D images, some works combine implicit neural representation with surface rendering techniques. 
These works typically require additional constraints for optimization, such as object masks. 
Moreover, inspired by the seminal work NeRF~\cite{wang2021nerf}, more recent works apply volume rendering techniques to optimize the implicit neural representation encoded geometry.
In this work, we adopt an implicit signed distance function as geometry representation and leverage the volume rendering technique proposed in NeuS~\cite{wang2021neus} to optimize the geometries with the ray-location correspondences~\cite{lyu2020differentiable,qian20163d,wu2018full}.

%% file: NeTO/sections/3_method.tex
\section{Method}
\label{sec:METHOD}
\subsection{Overview}
We aim to reconstruct the surfaces $\mathcal{S}$ of a solid transparent object from a set of posed object masks and the correspondences between the camera view rays and locations on the background under each viewpoint.
We propose to adopt an implicit Signed Distance Function (SDF) as surface representation and leverage volume rendering~\cite{wang2021neus} to enforce the refraction-tracing consistency, which enables stable and robust optimization.
Moreover, we propose a simple but effective strategy to identify the rays passing through self-occluded parts and then exclude these rays during optimization to avoid mistakenly enforcing refraction-tracing consistency. 

\begin{figure}[t]
\centering
\begin{overpic}
[width=1.0\linewidth]{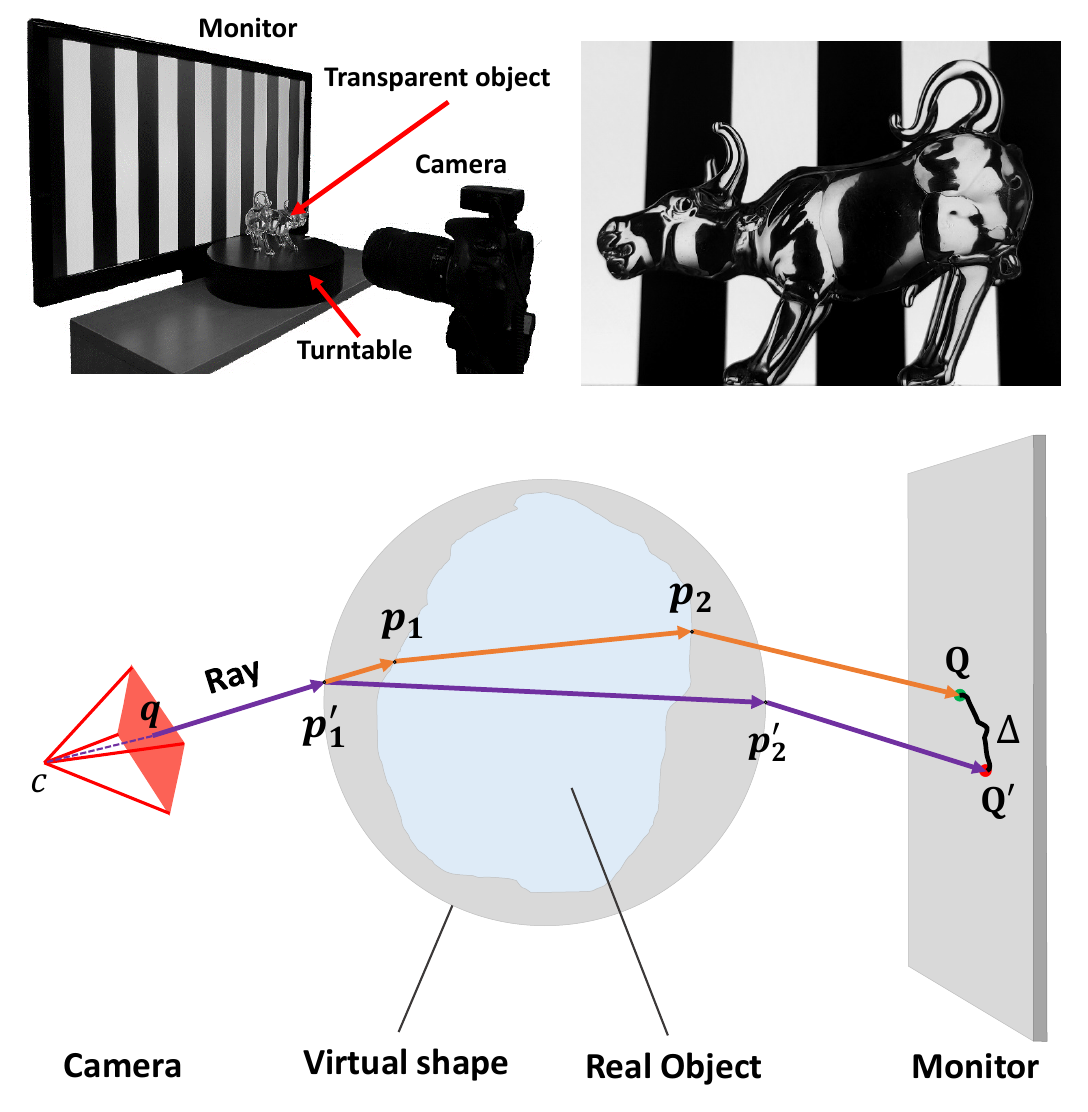}
\put(11, 61.5){ \small(a) Capture setup}
\put(22, -1.5){\small(c) Ray-location correspondence}
\put(60,61.5){ \small (b) Captured image}
\end{overpic}
\caption{
(a) Our transparent object capture setup;
(b) a captured image of a real Bull object;
(c) and the ray-location correspondence. (See details in preliminaries)}

\label{fig:ray_location}
\end{figure}

\subsection{Preliminaries}
 
    \textbf{Object capture setup.}
    To reconstruct the transparent objects, we adopt the similar object capture system proposed in ~\cite{lyu2020differentiable,wu2018full}.
    The system consists of a static LCD monitor, a turntable, and a camera.
    The monitor displays horizontal and vertical stripe patterns that form a Gray-coded background, and is placed behind the object and the camera.
    The transparent object is placed on the turntable, which is rotated in data acquisition to provide the static camera with multiple views of the object. 
    The silhouette mask information and environment matte can be extracted from the patterns displayed on the monitor.
 \begin{figure}[t]
    \centering
    \includegraphics[width=\linewidth,scale=1.00]{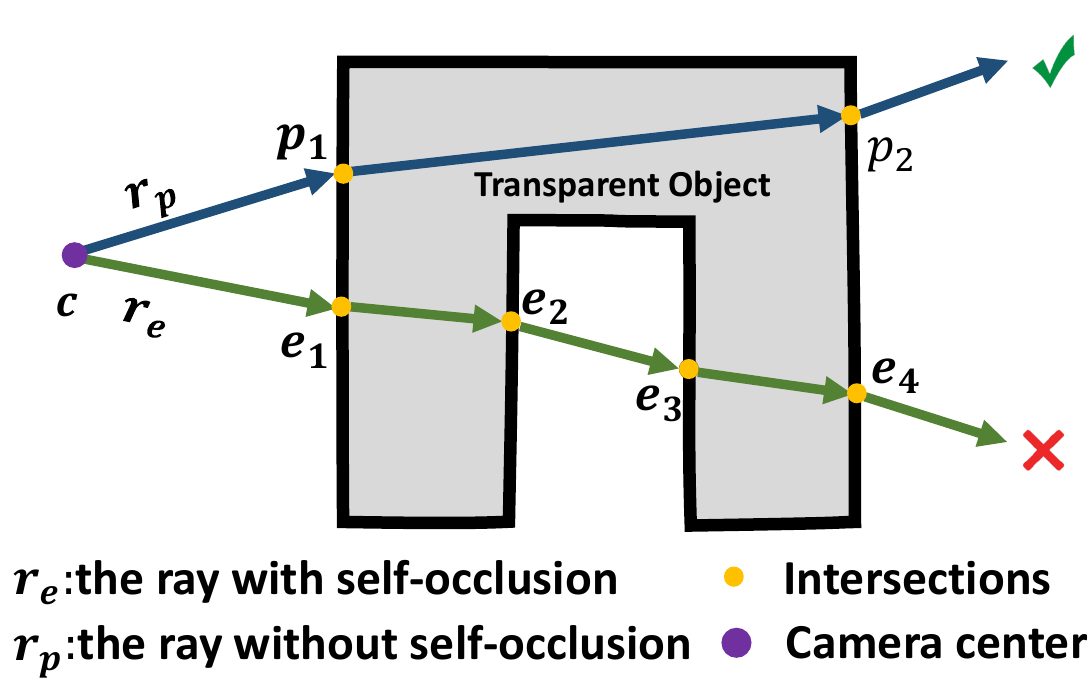}
    
    \caption{The diagram of a ray with self-occlusion $r_e$ and a ray without self-occlusion $r_p$. 
    The ray $r_p$ only refracts twice on the object surfaces, while the ray $r_e$ refracts on the surfaces more than twice due to self-occlusion.
    The rays with self-occlusion should be excluded in the optimization for high-quality reconstruction.
    }
    \label{fig:two_and_biger_two}
\end{figure}
    
    \textbf{Refraction-tracing consistency.}
    For general objects, feature points of the input images are extracted to establish correspondences for 3D reconstruction.
    However, for transparent objects, it's difficult to extract reliable feature points to establish correspondences, so the prior works and ours leverage the environment matting technique to establish the relationship between object geometry and the observed images.
    As shown in Figure~\ref{fig:ray_location}, a ray $r$ shooting from the camera center passes through the transparent object, which refracts twice on the object surfaces and then hits on the monitor at point $Q$.
    Since the gray-coded patterns are known, we can calculate the exact location of $Q$, and therefore we obtain a pair of camera ray $r$ and hit location $Q$.
    Our method is based on optimizing the correspondences between camera rays and the locations, which can also be named refraction-tracing consistency.

 \subsection{SDF-based refraction tracing}
    \textbf{Surface representation.}
    Unlike that the prior works adopt point clouds or meshes as geometry representations, we adopt Signed Distance Function (SDF) as surface representation.
    Specifically, the SDF field maps a point $x\in\mathbb{R}^3$ to its signed distance value to the surfaces, and the field is encoded by a Multi-layer Perceptrons (MLP) network. The surface $\mathcal{S}$ of the object is represented by the zero-set of the signed distance function (SDF), that is,\;$\mathcal{S} = \left\{ x \in  \mathbb{R}^3 | g(x) = 0\right\}$.
    The SDF field is initialized as a unit sphere. For convenience, we denote the shape being optimized as a virtual shape.

\textbf{Refraction-tracing.}
As shown in Figure~\ref{fig:ray_location}, given the current virtual shape, we first trace rays from the camera center that intersect and refract through the shape, and then optimize the SDF values of associated surface intersections according to the captured correspondences between the view rays and background locations (e.g., the ray $\overrightarrow{cq}$ and the location $Q$ in Figure~\ref{fig:ray_location}).
We take a ray that only refracts on the surfaces exactly twice as an example, the ray first enters the virtual shape at point $p_1^{\prime}$, and then it exits the shape at point $p_2^{\prime}$.
Finally, the simulated light path, shown in purple in Figure~\ref{fig:ray_location}, hits on the background monitor at a virtual location $Q^{\prime}$.
Before the optimization of geometry converges, $Q^{\prime}$ is generally different from the destination of the actual optical path passing through the real object, which is shown in orange, and finally hits on the background monitor at $Q$.
The goal of optimization is to minimize the differences between the virtual hitting location and real hitting location, that is, $\Delta=\left\|Q-Q^{\prime}\right\|^2$.

To trace how the simulated light path interacts with the virtual shape and then penalize the location differences in the optimization, we leverage the SDF-based volume rendering technique~\cite{wang2021neus} to calculate the exact locations of the two refraction intersections $p_1^{\prime}$ and $p_2^{\prime}$. The SDF-based surface rendering technique~\cite{yariv2020multiview} can also be used for the intersection calculation, as discussed in Section~\ref{surface_vs_Volume}, volume rendering yields more robust and stable optimization and leads to better reconstruction quality.

\begin{figure}[t]
\centering
\includegraphics[width=\linewidth,scale=1.00]{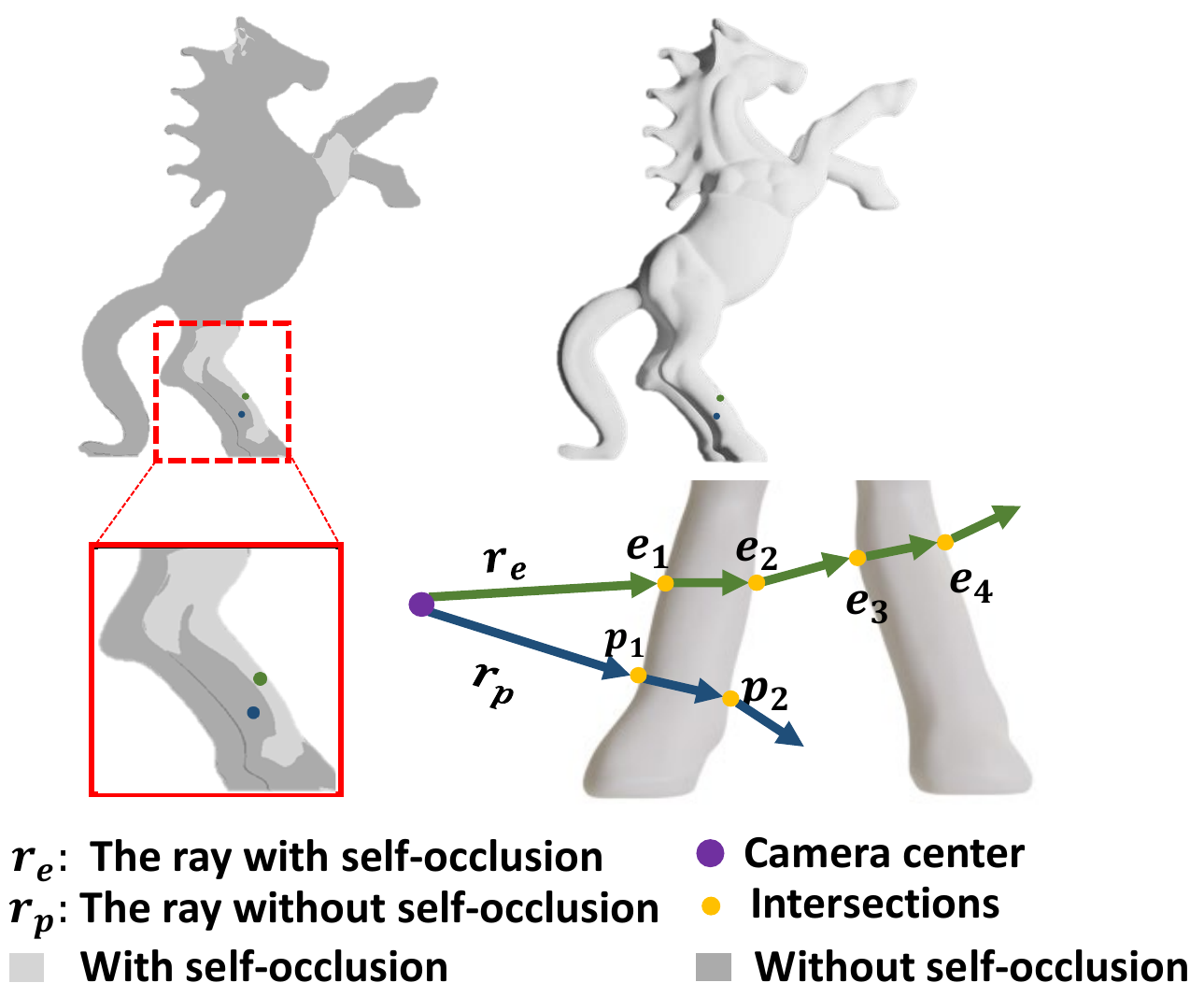}

\caption{A example of self-occlusion checking strategy applied on the real Horse object.}
\label{fig:self_occlusion_example}
\end{figure}

\subsection{Self-occlusion handling}
Since the objects to be reconstructed are solid, most camera rays will refract on the object surfaces exactly twice.
When a ray passes through self-occluded regions, the refractions will be more complex, and the ray may refract on the surfaces more than twice. 
As shown in Figure~\ref{fig:two_and_biger_two}, the light path without self-occlusion (blue line in Figure~\ref{fig:two_and_biger_two}) has two refracted intersections with object surfaces, while the light path with self-occlusion (green line) has four refracted intersections.
However, the prior works~\cite{lyu2020differentiable,wu2018full} ignore the self-occlusion problem and assume that all the camera rays only refract exactly twice. 
As a result, for the rays that refract more than twice, the simulated light paths will be mistakenly approximated, thus introducing wrong supervision information into the geometry optimization.

\textbf{Naive checking strategy.} To tackle this problem, the key is to identify whether a ray refracts more than twice, and then exclude the ray in the optimization.
A naive solution is to calculate the exact locations of the refraction intersections. As shown in Figure~\ref{fig:two_and_biger_two}, when a ray passes through the self-occluded parts, we can leverage Snell's law to obtain the directions of the refracted lights, and then calculate the locations of the intersections, $e_1, e_2, e_3, e_4$.
However, we have to extensively conduct iterative sampling and network queries to find the points sampled in the refracted lights which are in the surfaces, which significantly increases the computational costs.

\textbf{Proposed checking strategy.} 
We, therefore, propose a simple but effective strategy to identify the rays that refract more than twice at low costs.
The motivation is based on the {law of reversibility}, that is, \textit{If the direction of a light beam is reversed, it will follow the same path.}

\begin{figure}[t]
\centering
\includegraphics[width=\linewidth,scale=1.00]{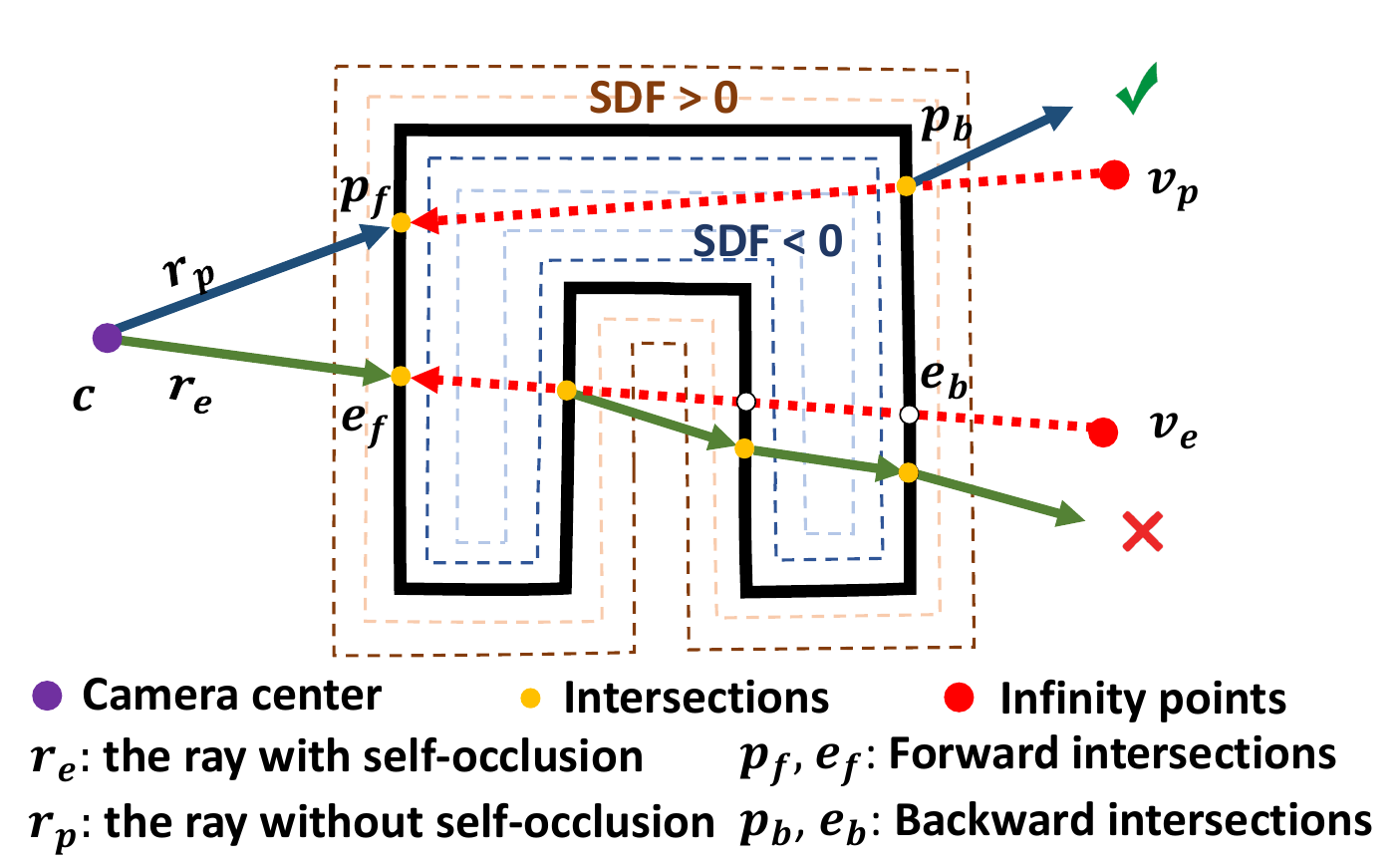}

\caption{The illustration of self-occlusion checking strategy. 
For the ray $r_p$, there are no surfaces on the line segment $\overline{ p_f p_b}$, where all SDF values of the sampled points are negative.
For the ray $r_e$, there exist surfaces on the line segment $\overline{ e_f e_b}$, where the SDF values of some sampled points are positive.
}
\label{fig:checking_strategy}
\end{figure}

As shown in Figure~\ref{fig:checking_strategy}, the procedure of the strategy is introduced below:

\begin{algorithm}

\caption{Self-occlusion checking strategy}
1) Shoot a ray $r_p$/$r_e$ emitting from the camera center, and get its first forward intersection $p_f$/$e_f$.

2) Leverage Snell's Law to obtain the refracted light line $\overrightarrow{p_f v_p}$/$\overrightarrow{e_f v_e}$, where $v_p$/$v_e$ is an infinite point on the line.

3) Shoot the reversed refracted light line $\overrightarrow{v_p p_f}$/$\overrightarrow{v_e e_f}$ from $v_p$/$v_e$, and then obtain the backward intersection $p_b$/$e_b$.

4) Sample points on the line segment $\overline{p_f p_b}$/$\overline{e_f e_b}$, and then evaluate the SDF values of the points.

5) If there exist points with positive SDF values, the ray refracts more than twice; if not, the ray refracts exactly twice.

\end{algorithm}

We can see the ray $r_p$ refracts the surfaces exactly twice, and there are no intersections on the line segment $\overline{p_f p_b}$, which indicates the light path $c \rightarrow p_f \rightarrow v_p$ is reversible.
On the other hand, for the ray $r_e$, there exist two more intersections on the line segment $\overline{e_f e_b}$, which indicates that the light path $c \rightarrow e_f \rightarrow v_e$ is not reversible with the twice refraction assumption.
Moreover, thanks to the properties of SDF (negative values inside and positive values outside), we can evaluate whether there exist any points with positive SDF values between the forward and backward intersections to identify the existence of self-occlusion.

Unlike that the naive checking strategy requires accurately finding the locations of all intersections, our proposed checking strategy only needs to identify whether there exist positive SDF values in a line segment with a short length.
We provide an example of the self-occlusion checking strategy applied on a real Horse object in Figure~\ref{fig:self_occlusion_example}, and our method can accurately identify the self-occluded regions (the overlapping legs of the horse). 
 




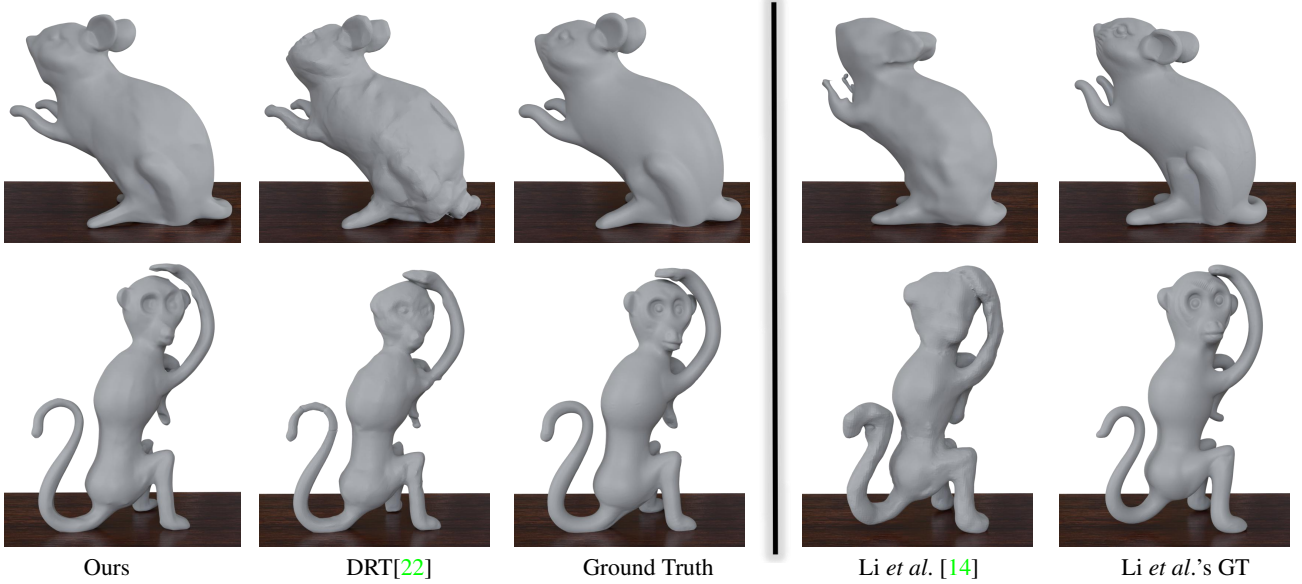
\begin{figure*}[tp]
\centering
\begin{overpic}
[width=1.0\linewidth]{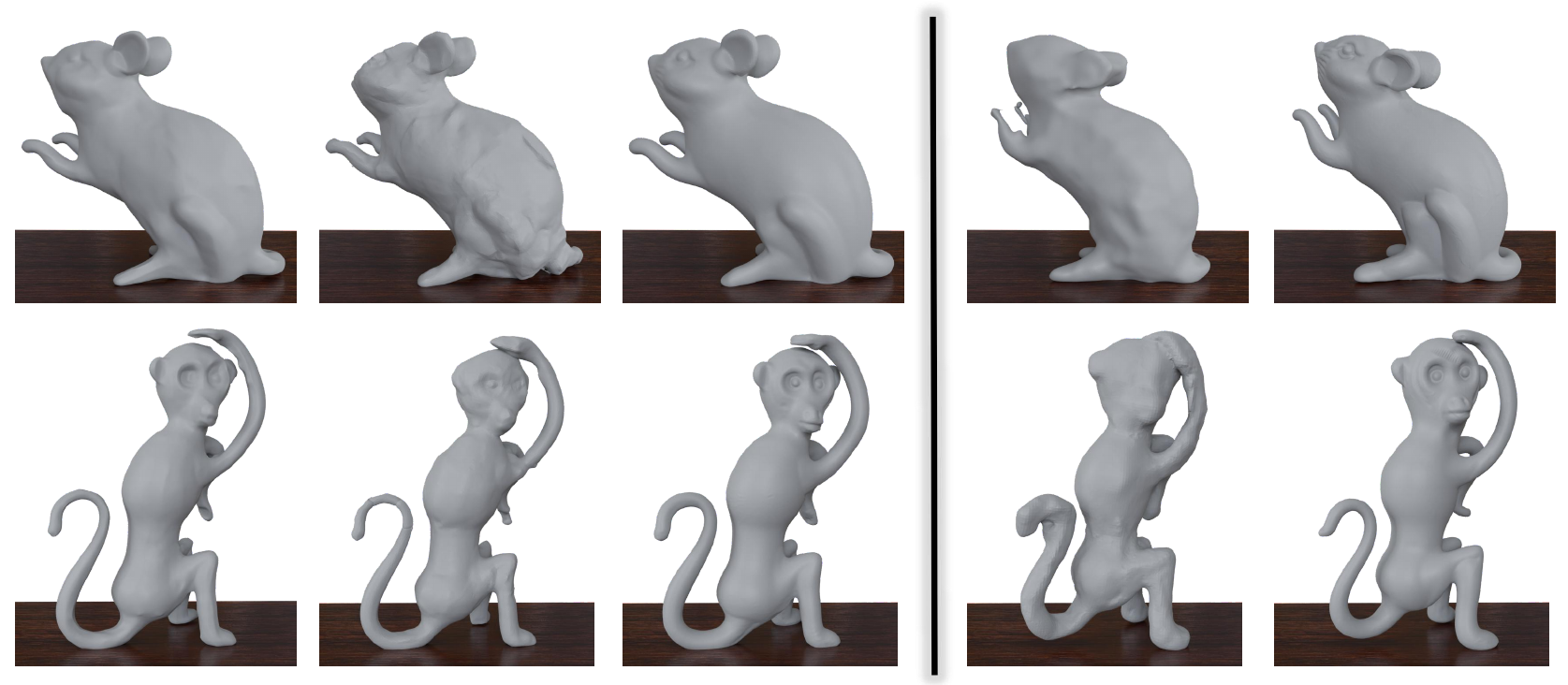}
\put(7,-1){\small Ours}
\put(27,-1){\small DRT\cite{lyu2020differentiable}}
\put(45,-1){\small Ground Truth}
\put(66,-1){\small Li~\etal\cite{li2020through}}
\put(86,-1){\small Li~\etal's GT}
\end{overpic}
\caption{Qualitative comparisons with $sparsity=8$ (9 views) on the Mouse and Monkey objects. Even with a limited set of images (9 images), Our method reconstructs faithful geometry with rich details.
However, DRT and Li~\etal\cite{li2020through} fail to reconstruct the geometries, the reconstructed models are over-smoothing, and the details are missing.
It should be noted that, due to different manufacturing batches, there are slight differences between the shapes used by Li~\etal and DRT, and therefore their results are compared to a different set of ground truth models.
}
\label{compare_with_four_views}
\end{figure*}

\subsection{Loss Functions}
We optimize the SDF field by sampling a batch of rays with their ray-location correspondences and object masks $\left\{Q, M\right\}$, where $Q$ is the observed location on the background monitor, and ${M}\in\left\{0, 1\right\}$ is mask value. 
We sample $n$ points on the ray, and the batch size is $m$. The loss function is defined as~$:$
\begin{equation}
\mathcal{L} = \omega_1\mathcal{L}_{Refraction} +  \omega_2\mathcal{L}_{Eikonal} + \omega_3\mathcal{L}_{Mask} 	
\end{equation}
	
\textbf{Refraction loss.} We minimize the difference between simulated background position $Q^{'}$ and and its corresponding captured
ground truth $Q$ (see Figure~\ref{fig:ray_location}). The refraction loss is defined as follows~$:$ 
\begin{equation}			
\mathcal{L}_{Refraction} = \sum_{i \in R}(\left\|Q_i-Q_i^{\prime} \right\|^{2})	
\end{equation}
where $R$ is the set containing ray paths that go through the object and refract on surfaces exactly twice.

With our proposed self-occlusion checking strategy, the invalid rays are excluded in the optimization, and the refraction loss gets rid of noises and makes reconstruction accurate.

 \begin{table}[t]
  \centering
  \resizebox{\linewidth}{!}{
    \begin{tabular}{ccccccc}
    \hline
        & \multicolumn{6}{c}{Sparsity=18 } 
        \\ \hline
          & \multicolumn{2}{c}{Li~\etal~\cite{li2020through}} & \multicolumn{2}{c}{DRT\cite{lyu2020differentiable}} & \multicolumn{2}{c}{\textbf{Ours}} 
          \\ \hline
          
          & Acc. / Com.$\downarrow$ & F-score$\uparrow$
          & Acc. / Com.$\downarrow$ & F-score$\uparrow$ 
          & Acc. / Com.$\downarrow$& F-score$\uparrow$  
          \\ \hline
    Pig   & 2.63 / 3.01 & 0.18  & 1.87 / 1.45 & 0.35  & \textbf{0.91 / 0.88} & \textbf{0.47 } \\
    Dog   & 3.27 / 2.87 & 0.22  & 1.51 / 1.39 & 0.36  & \textbf{0.88 / 0.78} & \textbf{0.57 } \\
    Mouse & 1.93 / 2.80 & 0.25  & 2.90 / 2.29 & 0.23  & \textbf{1.27 / 1.09} & \textbf{0.41 } \\
    Monkey & 2.59 / 3.02 & 0.20  & 2.56 / 1.60 & 0.23  & \textbf{1.02 / 0.91} & \textbf{0.41 } \\
    Horse &   /    &   /    & 1.95 / 1.08 & 0.51  & \textbf{0.86 / 0.79} & \textbf{0.66 } \\
    Tiger &   /    &    /   & 3.04 / 1.74 & 0.40  & \textbf{1.01 / 0.86} & \textbf{0.59 } \\
    Rabbit &   /    &  /     & 1.44 / 1.27 & 0.38  & \textbf{1.07 / 0.93} & \textbf{0.50 } \\
    Hand  &    /   &   /    & 1.32 / 1.49 & 0.19  & \textbf{0.60 / 0.53} & \textbf{0.63 } \\
     \hline
    Avg.  & 2.60 / 2.92 & 0.21  & 2.07 / 1.53 & 0.33  & \textbf{0.95 / 0.84} & \textbf{0.53 } \\
     \hline
    \end{tabular}%
    }
    \caption{Evaluation of reconstruction with $sparsity=18$ (4 views).  
    Compared with Li~\etal~\cite{li2020through} and DRT~\cite{lyu2020differentiable}, our method achieves the best performance in all cases.
} 
\label{table:compare1}
\end{table}%

\textbf{Mask loss.} Following the prior works~\cite{lyu2020differentiable, wu2018full}, the mask loss is also included and defined as~$:$
\begin{equation}
    \mathcal{L}_{mask} = BCE(M_k, O_k) \label{mask_loss}
\end{equation}
where $O_k$ is the sum of weights along the $k_{th}$ camera ray, $M_k$ is the mask of the $k_{th}$ ray, and $BCE$ is the binary cross entropy loss.

\textbf{Eikonal loss.} We add an Eikonal loss to regularize the SDF field of the sampling point on the ray to have a unit norm of gradients. The loss term is defined as~$:$
\begin{equation}
\mathcal{L}_{Eikonal} = \frac{1}{nm}\sum_{k,i}(||\triangledown g({x}_{k,i})||_2 - 1) ^2 \label{Eikonal_loss}
\end{equation}
 where ${x}_{k,i}$ is the $i_{th}$ sampled point at the $k_{th}$ ray, $g$ is the geometry function.

\begin{table}[t]
  \centering
  \resizebox{\linewidth}{!}{
    \begin{tabular}{ccccccc}
    \hline
        & \multicolumn{6}{c}{Sparsity=9}      
        \\ \hline
          & \multicolumn{2}{c}{Li~\etal~\cite{li2020through}} & \multicolumn{2}{c}{DRT\cite{lyu2020differentiable}} & \multicolumn{2}{c}{\textbf{Ours}}  
          \\ \hline        
          & Acc. / Com.$\downarrow$ & F-score$\uparrow$
          & Acc. / Com.$\downarrow$ & F-score$\uparrow$ 
          & Acc. / Com.$\downarrow$& F-score$\uparrow$  
          \\ \hline
    Pig         
    & 1.56 / 1.77 & 0.23  & 0.90 / 0.91 & 0.56  & \textbf{0.83 / 0.77} & \textbf{0.60 } \\
    Dog   
    
    & 1.15 / 1.19 & 0.41  & 1.48 / 1.49 & 0.33  & \textbf{0.83 / 0.74} & \textbf{0.58 } \\
    Mouse 
    & 1.54 / 1.63 & 0.29  & 1.32 / 1.52 & 0.36  & \textbf{0.79 / 0.72} & \textbf{0.50 } \\
    Monkey 
    & 1.61 / 1.52 & 0.25  & 1.18 / 1.27 & 0.30  & \textbf{0.88 / 0.80} & \textbf{0.42 } \\
    Horse 
    &   /    &   /    & 0.68 / 0.60 & 0.80  & \textbf{0.68 / 0.45} & \textbf{0.81 } \\
    Tiger 
    &   /    &   /    & 1.58 / 1.23 & 0.59  & \textbf{0.85 / 0.69} & \textbf{0.71 } \\
    Rabbit 
    &   /    &   /    & 0.75 / 0.77 & 0.62  & \textbf{0.67 / 0.57} & \textbf{0.73 } \\

    Hand  
    &    /   &   /    & 0.88 / 0.98 & 0.32  & \textbf{0.60 / 0.53} & \textbf{0.63 } \\
     \hline
    Avg.  
    & 1.46 / 1.52 & 0.29  & 1.09 / 1.09 & 0.48  & \textbf{0.76 / 0.65} & \textbf{0.62 } \\
     \hline
    \end{tabular}%
    }
    \caption{Evaluation of reconstruction with $sparsity=8$ (9 views).  
    Compared with Li~\etal~\cite{li2020through} and DRT~\cite{lyu2020differentiable}, our method achieves the best performance in all cases.
} 
\label{table:compare2}
\end{table}%

%% file: NeTO/sections/4_implementation.tex
\section{Experiments}

\subsection{Experimental Setup}
\textbf{Datasets.} We conduct evaluations on the DRT~\cite{lyu2020differentiable} dataset. The dataset contains eight transparent objects. Each transparent object contains 72 views with corresponding masks, ray-pixel correspondences, and extrinsic and intrinsic camera parameters. 
The view resolution is $960\times1280$ or $1080\times1920$.
Ground truth 3D models are also provided for the transparent models. 
\begin{table}[t]
  \centering
  \resizebox{\linewidth}{!}{
    \begin{tabular}{ccccccc}
    \hline
        & \multicolumn{6}{c}{Sparsity=4}      
        \\ \hline
          & \multicolumn{2}{c}{Li~\etal~\cite{li2020through}} & \multicolumn{2}{c}{DRT\cite{lyu2020differentiable}} & \multicolumn{2}{c}{\textbf{Ours}}  
          \\ \hline        
          & Acc. / Com.$\downarrow$ & F-score$\uparrow$
          & Acc. / Com.$\downarrow$ & F-score$\uparrow$ 
          & Acc. / Com.$\downarrow$& F-score$\uparrow$  
          \\ \hline

    Pig &0.90 / 1.14 & 0.47  & 0.73 / 0.75 & 0.65  & \textbf{0.70 / 0.64} & \textbf{0.65 } \\
    Dog &0.83 / 0.95 & 0.53  & 1.01 / 1.02 & 0.48  & \textbf{0.80 / 0.71} & \textbf{0.61 } \\
    Mouse &1.68 / 1.69 & 0.28  & 1.16 / 1.39 & 0.40  & \textbf{0.78 / 0.70} & \textbf{0.50 } \\
    Monkey &1.32 / 1.24 & 0.34  & 1.03 / 1.10 & 0.37  & \textbf{0.86 / 0.78} & \textbf{0.43 } \\
    Horse   &  / &   /    & 0.65 / 0.57 & 0.83  & \textbf{0.66 / 0.47} & \textbf{0.85 } \\
     Tiger  & /   &    /   & 1.01 / 0.87 & 0.68  & \textbf{0.73 / 0.58} & \textbf{0.79 } \\
     Ribbit & /     & /      & 0.69 / 0.72 & 0.66  & \textbf{0.61 / 0.51} & \textbf{0.79 } \\
     Hand  & /   &   /    & 0.84 / 1.01 & 0.43  & \textbf{0.63 / 0.49} & \textbf{0.71 } \\ \hline
    Avg. &1.18 / 1.25 & 0.40  & 0.89 / 0.92 & 0.56  & \textbf{0.72 / 0.61} & \textbf{0.66 } \\
     \hline
    \end{tabular}%
    }
    \caption{Evaluation of reconstruction with $sparsity=4$ (18 views).  
    Compared with Li~\etal~\cite{li2020through} and DRT~\cite{lyu2020differentiable}, our method achieves the best performance in all cases.
} 
\label{table:compare3}
\end{table}%
\begin{figure}[t]
\centering
\begin{overpic}
[width=1.0\linewidth]{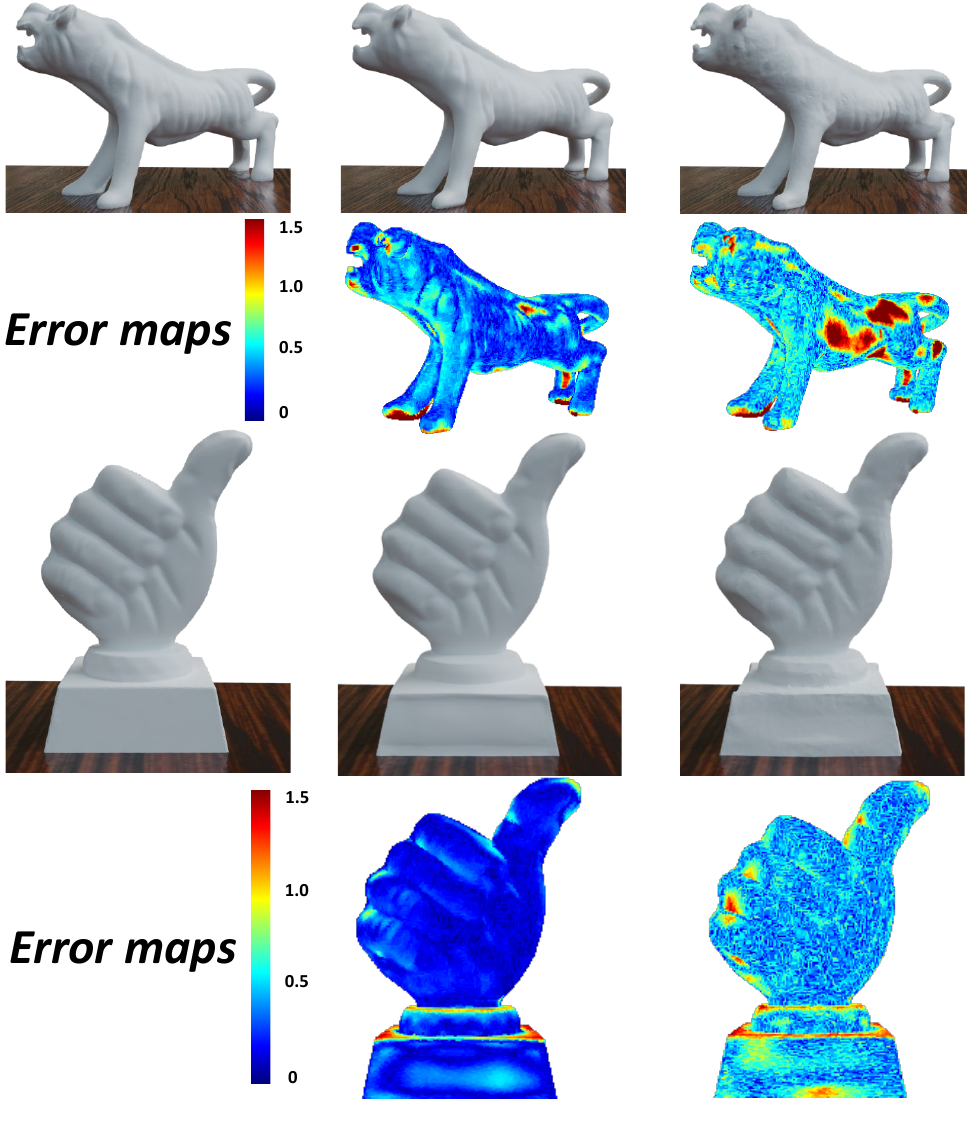}
\put(2.5, -2.4){Ground Truth}
\put(37,-2.4){Ours}
\put(66,-2.4){DRT\cite{lyu2020differentiable}}
\end{overpic}
\caption{We show two groups of full views reconstruction results generated by ours and DRT~\cite{lyu2020differentiable}, respectively. Our method can faithfully reconstruct high-quality geometries with fewer errors.}
\label{figure-additiaonal-full}
\end{figure}

\textbf{Implementation Details.} The geometry function $g$ is modeled by an MLP, which consists of 8 hidden layers with a hidden size of 256. We use PyTorch ~\cite{paszke2017automatic} to implement our approach and use the Adam optimizer with a global learning rate $5e^{-4}$ for the network training. Our network architecture and initialization scheme are similar to those of prior works~\cite{wang2021neus,wang2022neuralroom}. We sample $512$ rays per batch and train our model for 300k iterations on a single NVIDIA RTX 2080Ti GPU. We extract explicit mesh from the learned SDF field via a marching cube algorithm~\cite{lorensen1987marching}.

A hierarchical sampling strategy is used to sample points along a ray in a coarse-to-fine manner for volume rendering. We first uniformly sample 64 points along the ray, and then iteratively conduct importance sampling~\cite{wang2021neus} to sample more points on top of coarse probability estimation for $4$ times. The positional encoding is applied to the spatial location with 5 frequencies. 
The hyper-parameters used in the experiments are set as $\omega_1=0.0001, \omega_2=0.1, \omega_3=0.1$. 
Following the prior work~\cite{wu2018full}, the IOR (index of refraction) of air is set to 1.0003 and the IOR of transparent material (glass) is set to 1.4723.

%% file: NeTO/sections/5_experiments.tex
\subsection{Comparisons}

\textbf{Baselines.} 
We compare our method with the two state-of-the-art baselines: 1) DRT~\cite{lyu2020differentiable}, the most related work to ours, which also optimizes the geometry by the ray-location correspondences but it adopts explicit mesh as surface representation.
2) A data-driven deep learning based approach Li et al. [2020]~\cite{li2020through}. They generate a synthetic dataset of the transparent objects, and then learn geometric priors from the training data to reconstruct the objects.

 \begin{table}[t]
\centering
\resizebox{\linewidth}{!}{
\begin{tabular}{c|ccc|ccc}
\hline

\multicolumn{1}{c}{}  
  		
& \multicolumn{3}{c}{DRT\cite{lyu2020differentiable}} 
& \multicolumn{3}{c}{Ours}\\ \cline{2-7}
\hline
& Acc $\downarrow$  & Comp $\downarrow$   & F-score $\uparrow$  & Acc $\downarrow$  & Comp $\downarrow$ & F-score $\uparrow$ \\
\hline
 Pig     & 0.6566 & 0.6863 & 0.7142
 & \textbf{0.5669} & \textbf{0.4689} & \textbf{0.8474} \\
Dog     & 0.9072 & 0.8704  &0.5526
    &\textbf{0.7601} &\textbf{0.6274} 
    &\textbf{0.7029} \\
Mouse   & 0.8018 & 0.839 & 0.5226 
        &\textbf{0.7788} 
        &\textbf{0.6811}  
        &\textbf{0.5998}\\
Monkey  & 0.945 & 0.8923  & 0.4422
& \textbf{0.8415} & \textbf{0.7467} & \textbf{0.4827} \\

Horse  & 0.6636 & 0.6095 & 0.8422 &\textbf{0.6193} & \textbf{0.4099} 
&\textbf{0.884}  \\
Tiger & 0.8191 & 0.723 & 0.7665 &\textbf{0.7099} & \textbf{0.5705} & \textbf{0.7979}  \\
Rabbit  & 0.5971 & 0.6202  & 0.7686 
& \textbf{0.5839} & \textbf{0.4941} & \textbf{0.8324} \\
 Hand  & 0.4792 & 0.5856  & 0.5796
 & \textbf{0.3947} & \textbf{0.3140}  & \textbf{0.7740}  \\ \hline
 Avg. 
 & 0.7337 & 0.7282 
 & 0.6485
 & \textbf{0.6568} 
 & \textbf{0.5390} 
 & \textbf{0.7401} \\ \hline
\end{tabular}
}

\caption{Comparisons of reconstruction with full views. 
Our method obtains the best performance in all cases, and the full table is in the supplementary material.
}

\label{table:compare_full_views}
\end{table}
\begin{figure}[t]

\centering
\begin{overpic}
[width=1.0\linewidth]{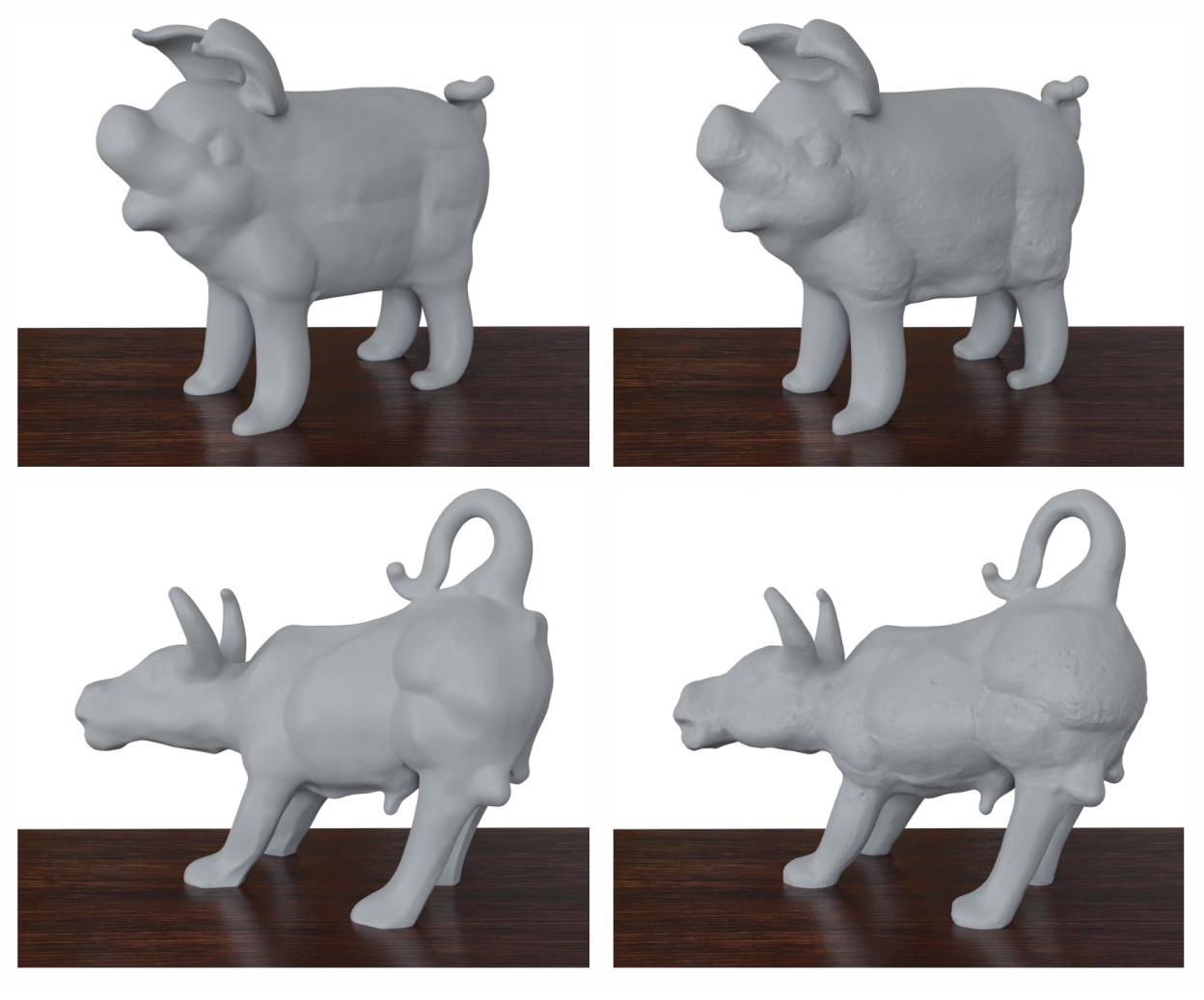}
\put(21, -2){Ours}
\put(69, -2){\small DRT\cite{lyu2020differentiable}}
\end{overpic}
\caption{Comparisons on our self-collected data. Our method reconstructs high-quality surfaces, while the surfaces recovered by DRT contain lots of noise.}
\label{figure:real_object}

\end{figure}
\begin{figure*}[htbp]
\centering
\begin{overpic}
[width=1.0\linewidth]{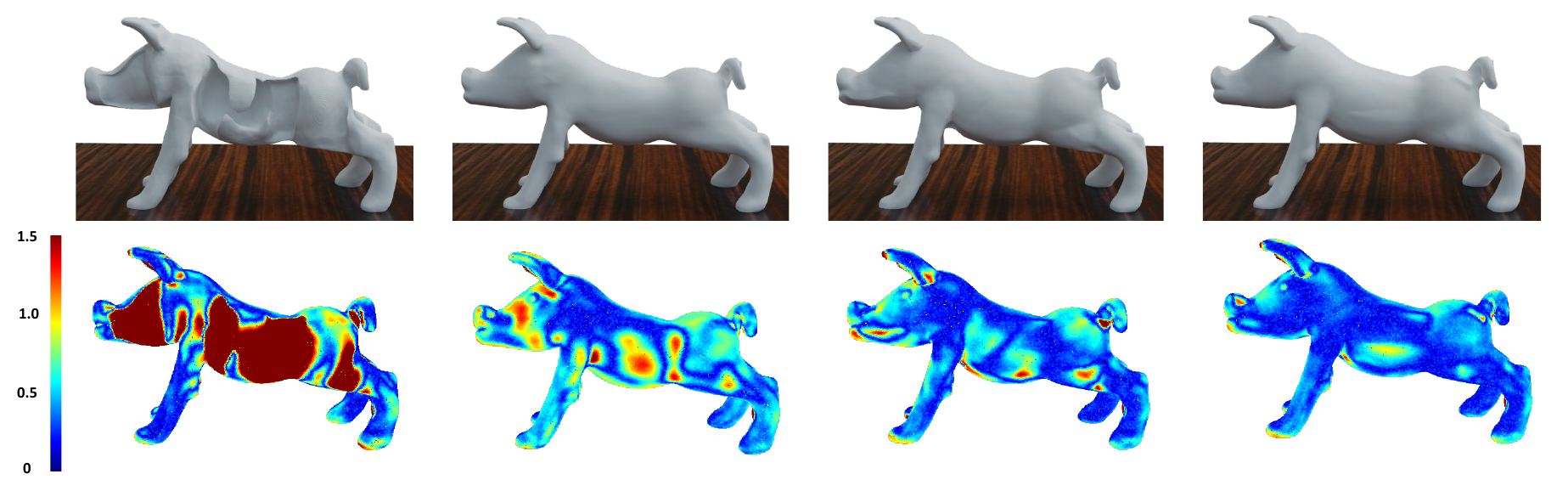}
\put(8, -1){\small W/o $\mathcal{L}_{Eikonal}$}
 \put(33.5, -1){\small W/o $\mathcal{L}_{Refraction}$}
 \put(59, -1){\small W/o Self-occlusion}
\put(88,-1){Full}

\end{overpic}
\caption{Qualitative ablation study on the Pig model.
For better visualization, we measure and colorize the errors between the reconstructed models with the ground truth model.
The reds indicate large errors, and the blues indicate small errors.
}
\label{Ablation}
\vspace{-3mm}
\end{figure*}
\textbf{Evaluation Protocols.} 
To evaluate the quality of reconstructed models, we calculate the metrics, accuracy, completeness, precision, recall, and F-score between the reconstructed model and the ground truth model. 
Our method and DRT adopt the same dataset provided by DRT, so the input images and the ground truth models adopted by ours and DRT are the same.
Although Li et al. experimented with transparent objects obtained from the same source, due to different manufacturing batches, there are slight differences between the shapes, and therefore their results are compared to a different set of ground truth models.
For fairness, the ground truth models of the two types are reshaped into the same scale for evaluation.
We evaluate the reconstruction results with sparse views and with full dense views.

\textbf{Reconstructions with sparse views.} 
We conduct experiments under various sparsity levels. The sparse setting selects a small proportion of the camera views
by consecutively sampling a view $I_i$ from every $sparsity = n$
camera index, i.e., $\{1, n+1, 2n+1, ... \}$.
The quantitative comparisons with $sparsity=18, 8, 4$ are presented in Table~\ref{table:compare1}, Table~\ref{table:compare2} and Table~\ref{table:compare3} respectively.
As you can see, with sparse views, our results outperform DRT and Li~\etal in terms of model completeness and accuracy. 
In addition to making quantitative comparisons, we render the model to visually observe the differences between our method and other methods. 
The qualitative comparisons are shown in Figure~\ref{compare_with_four_views}, and our method faithfully reconstructs the geometry with rich details, such as the leg of the Mouse and the eyes of the Monkey.
The reconstruction results produced by~\cite{li2020through} and~\cite{lyu2020differentiable} do not work well to reconstruct the rich geometric details and tend to over-smooth the surfaces. 
	
\textbf{Reconstruction with full views.}
When we use more views, e.g., full views (72 views), our reconstruction results and DRT are improved compared with reconstructions with sparse views.
However, based on the quantitative results presented in Figure~\ref{figure-additiaonal-full} and the qualitative results shown in Table~\ref{table:compare_full_views}, our method significantly outperforms the other methods in terms of completeness and accuracy, and the reconstructed models contain more rich details and have fewer errors.
We further evaluate a self-collected real Bull and Pig object, as shown in Figure~\ref{figure:real_object}. Our method accurately recovers the geometry with clean and smooth surfaces, while DRT mistakenly reconstructs surfaces with noises..

\begin{table}[t]
\centering
\resizebox{\linewidth}{!}{
\begin{tabular}{lccccc}	
    \hline
    
    Method &Acc $\downarrow$  &Comp $\downarrow$     &Recall $\uparrow$  &Prec $\uparrow$    &F-score $\uparrow$ \\ \hline
    
    W/o $\mathcal{L}_{Eikonal}$     & 3.3086 & 1.5212 & 0.4000   & 0.5384 & 0.4597 \\
    W/o $\mathcal{L}_{Refraction}$     & 0.7579 & 0.7019 & 0.5900  & 0.6452 & 0.6180 \\
    W/o Self-occlusion     & 0.6530 & 0.5440 & 0.7319  & 0.7886 & 0.7592 \\
    full  & \textbf{0.5669} & \textbf{0.4689} & \textbf{0.8300} & \textbf{0.8670} & \textbf{0.8474} \\

    \hline
\end{tabular}
}
\caption{Ablation study on Pig model. We test the effect of the Eikonal loss, refraction loss, and self-occlusion strategy used in the method. 
This analysis shows that the Full performs the best quantitatively.
}
\label{table:a}	
\end{table} 

\subsection{Ablation study and discussions.}

 \textbf{Ablation study.}
To better validate the effects of the self-occlusion checking strategy and the optimization terms, we conduct the ablation studies on the full method, method without self-occlusion checking, method without Eikonal loss term, and method without refraction loss term.
The quantitative evaluation is shown in Table~\ref{table:a}, and the qualitative evaluation is presented in Figure~\ref{Ablation}.
The experiments demonstrate that $\mathcal{L}_{Eikonal}$ plays the most important role, which encourages the SDF field to be continuous and smooth.
Without the refraction loss term, although our method can still reconstruct the rough shape relying on the silhouette information, the reconstruction becomes worse with larger errors.
Thanks to our proposed self-occlusion checking strategy, the quality of the self-occluded parts is improved with fewer errors, like the legs of the Pig model. 
\begin{figure}[h]
    \centering
    \setlength{\abovecaptionskip}{0cm}
    \setlength{\belowcaptionskip}{0cm}
 \captionsetup{font={small},justification=raggedright}
		\begin{overpic}
          [width=1.0\linewidth]{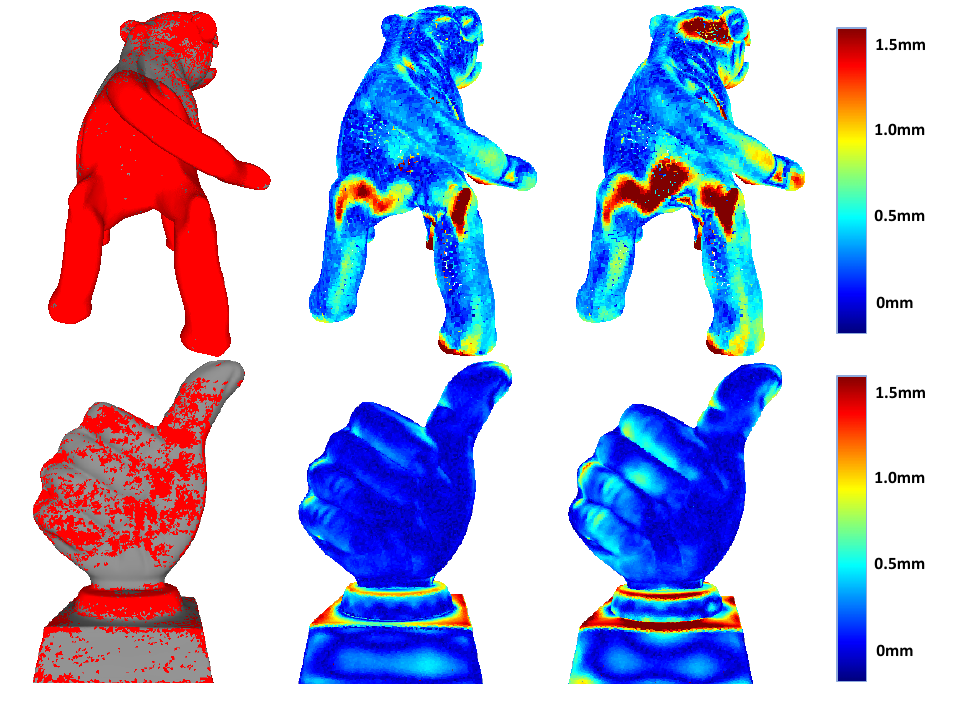}
          \put(0.5,-3){ Affected area}
          \put(31.5,-3){ W/ check}
          \put(59.5,-3){ W/o check}
        \end{overpic}
        \vspace{0.05cm}
        
        \caption{W/ and W/o Self-occluded check with full views.}
		\label{figure:tiger_hand}
\vspace{-3mm} 
\end{figure}
Furthermore, to better demonstrate the effectiveness of the self-occlusion checking strategy, we visualize the reconstruction error maps and the object surfaces affected by all self-occluded rays in Figure~\ref{figure:tiger_hand} (a self-occluded ray has multiple refractions on object surfaces, and thus a few rays will cause more affected surfaces), where the geometries are improved with self-occlusion check, especially on the affected regions.

\textbf{Different rendering techniques.} 
\label{surface_vs_Volume}
Both surface rendering~\cite{niemeyer2020differentiable} and volume rendering~\cite{wang2021neus} are used in neural rendering-based reconstruction.
Through experiments, we find that optimization using volume rendering is more robust and stable than using surface rendering.
As shown in Figure~\ref{rendering}, the reconstructed model using surface rendering is over-smoothing and lacks detailed geometries, while the reconstruction model using volume rendering achieves much better quality with rich details.

%% file: NeTO/sections/6_conclusion.tex
\section{Conclusion and Future Work}
We have presented NeTO, a novel neural rendering-based method for transparent object reconstruction, which adopts implicit signed distance function as surface representation and leverage volume rendering to enforce refraction-tracing consistency.
With our proposed self-occlusion checking strategy, the reconstructed geometries of self-occluded parts are further improved.
Our method significantly outperforms the state-of-the-art methods qualitatively and quantitatively by a large margin.

Although our method achieves high-quality reconstruction of transparent objects, the objects should be solid. 
This is because we adopt the ray-location correspondences, which assumes that most of the camera rays only refract on the object surfaces exactly twice.
In the future, we would like to explore how to reconstruct hollow transparent objects, where refraction is more complex and
most of the camera rays will refract on the surfaces more than twice.

\begin{figure}[t]
\centering
\begin{overpic}
[width=1.0\linewidth]{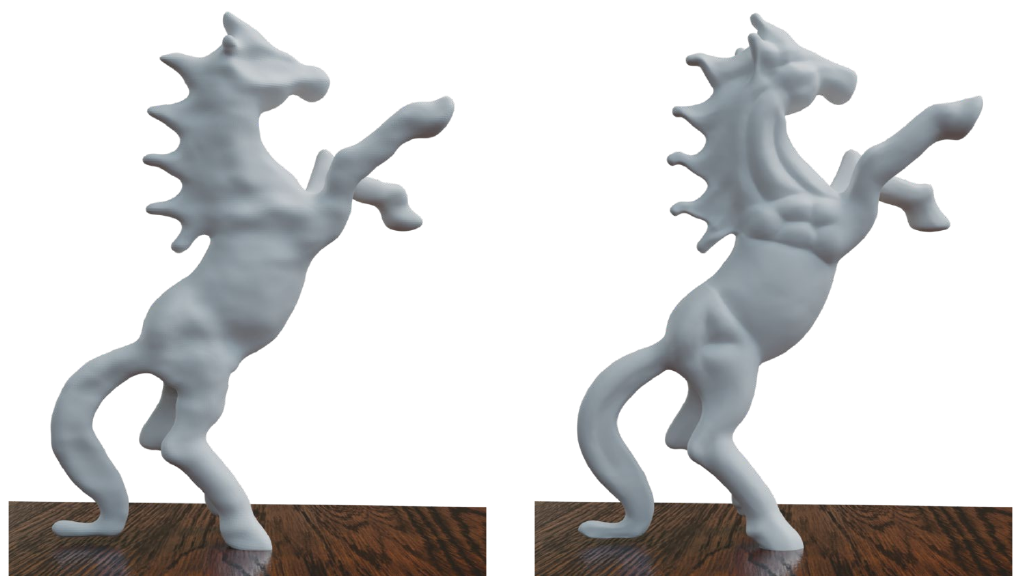}
\put(8.5, -2.5){Surface rendering}
\put(60.5,-2.5){Volume rendering}
\end{overpic}
\caption{Reconstruction with volume or surface rendering. The F-score values (from left to right) are 0.472 and 0.884, respectively.}
\label{rendering}
\end{figure}
\section{Acknowledgments}
This work is partially supported by NSFC (No.61972298).